\pdfoutput=1

\documentclass[11pt]{article}

\usepackage[]{acl}

\usepackage{times}
\usepackage{latexsym}
\usepackage{graphicx}

\usepackage[T1]{fontenc}

\usepackage[utf8]{inputenc}

\usepackage{microtype}

\title{Attention at SemEval-2023 Task 10: Explainable Detection of Online Sexism (EDOS)}

\author{Debashish Roy, \ Manish Shrivastava\\
IIIT Hyderabad, India \\
\tt debashish.roy@students.iiit.ac.in \\
\tt m.shrivastava@iiit.ac.in\\
}
    
\begin{document}
\maketitle
\begin{abstract}
 In this paper, we have worked on interpretability, trust, and understanding of the decisions made by models in the form of classification tasks. The task is divided into 3 subtasks. The first task consists of determining Binary Sexism Detection. The second task describes the Category of Sexism. The third task describes a more Fine-grained Category of Sexism. Our work explores solving these tasks as a classification problem by fine-tuning transformer-based architecture. We have performed several experiments with our architecture, including combining multiple transformers, using domain adaptive pretraining on the unlabelled dataset provided by Reddit and Gab, Joint learning, and taking different layers of transformers as input to a classification head. Our system (with team name ‘Attention’) was able to achieve a macro F1 score of 0.839 for task A, 0.5835 macro F1 score for task B and 0.3356 macro F1 score for task C at the Codalab SemEval Competition. Later we improved the accuracy of Task B to 0.6228 and Task C to 0.3693 in the test set.
\end{abstract}

\section{Introduction}

Online sexism is a growing problem in this era, considering the large scale of online texts. 
Detecting online Sexism is an important NLP task, and has a lot of social impacts related to it. Sexism is defined as any abuse or negative sentiment that is directed towards women based on their gender or based on their gender combined with one or more other identity attributes (e.g. black women, trans women). 

Automated tools are widely available to detect sexism but mostly don’t work with the issue of explaining why the text is sexist. Flagging what is sexist content and also explaining why it is sexist is very important aspects of these automated tools. We tried to focus on interpretability, trust, and understanding of the decisions made by models. Having an automated system that is capable of understanding, interpreting, and classifying sexist text will be a major step to make online sites safe. 

Explainable Detection of Online Sexism (EDOS) at SemEval 2023 \cite {edos2023semeval}, tries to solve this problem by exploring different systems that can categorize the dataset created with texts from Reddit and Gab. 

The problem is divided into 3 subtasks. The first task is formulated as Binary Sexism Detection, where the main idea is to detect whether a text is sexist or not. 

The second task describes the Category of Sexism. So, if the text is sexist, under which category it belongs? The categories are threats, derogation (treating someone as little worthy), animosity (a strong dislike or unfriendly feeling), and prejudiced discussions (an unfair feeling of dislike for a person or group because of race, sex, religion, etc). This makes this task a four-class classification problem given the text is sexist.  

The third task describes the Fine-grained Vector of Sexist text. For posts that are sexist, systems were supposed to predict more fine-grained aspects of the category of sexism. Which can help in interpretability and understanding why the post is sexist on a much deeper level. The sexist posts are classified into an 11-class classification problem, given the text is sexist. The 11 classes are Descriptive attacks, Aggressive and emotive attacks, Casual use of gendered slurs, Profanities, and insults, Immutable gender differences and gender stereotypes, Supporting systemic discrimination against women as a group, Incitement, and encouragement of harm, Dehumanising attacks \& overt sexual objectification, Supporting mistreatment of individual women, Backhanded gendered compliments, Threats of harm, Condescending explanations or unwelcome advice.

To solve this problem, we have fine-tuned transformer-based architecture. Our final submission is made up of two models: RoBERTa \cite{zhuang-etal-2021-robustly} and DeBERTa \cite{https://doi.org/10.48550/arxiv.2006.03654}. The last layer is then passed to the MLP. At the last layer, we have the number of classes. We have also explored Domain Adaptive Pre Training using masked language modeling on both of these transformers and used their weights for the classification task. The models remain the same for all the tasks except the last layer changes based on the number of classes. 


Our system ranked 35\textsuperscript{th} in sub-task 1, 50\textsuperscript{th} in sub-task 2, and 44\textsuperscript{th} in sub-task 3, out of 583 participants. 

All of our code is made publicly available
on Github\footnote{\url{https://github.com/debashish05/Explainable_Detection_of_Online_Sexism}}. The domain adaptive pretrained versions of RoBERTa and DeBERTa are also made publicly available at Huggingface 
\footnote{\url{https://huggingface.co/debashish-roy/}}
 
\section{Background}

Transformer-based models such as RoBERTa and DeBERTa has shown outstanding performance on wide domains of NLP, including text classification. These models can capture a sort of structure in the hateful speech as well \cite{basile-etal-2019-semeval}. 

Domain Adaptive Pre Training is used in many domains of NLP, to make the model focus on the specific domain rather than a large variety of text it is trained on. Don't Stop Pre Training: Adapt Language Models to Domains and Tasks \cite{https://doi.org/10.48550/arxiv.2004.10964} talk about the increase in the performance of a model if it is pretrained on similar types of data.  

SemEval-2021 Task 7: HaHackathon, Detecting and Rating Humor and Offense \cite{meaney-etal-2021-semeval} talks about the importance of pretraining and using ensembling methods to achieve high accuracy. It also described the importance of lexical features.

This SemEval task itself provided the dataset. The dataset contains text in English. The dataset contains 20,000 texts. Half of the text was taken from Gab and the other half from Reddit. The entries are then manually annotated by three trained annotator women. For each text, we have (a) label-sexist, describing whether the label is sexist or not, (b) label-category: given the text is sexist what is the category of sexism, in the case of non-sexist text it is none. (c) label-vector: it is more fine-grained details about the category of sexism. 

The data is divided into 70\% as training data, 10\% as validation data, and 20\% as test data. The distribution of labels for all three tasks is described in Table 1, Table 2, and Table 3. The major issue with the dataset for task C is less number of samples. All of the text is less than 64 words, so while generating tokens for transformers we have kept the max length to 64.

\begin{table}[h]
\small
    \centering
    \begin{tabular}{|c|c|}
        \hline
        \textbf{Classes} & \textbf{Instances}\\
        \hline
        Sexist Text & 3398 \\
        \hline
        Non-Sexist Text & 10602\\
        \hline
    \end{tabular}
    \caption{Class distribution for sub-task 1 train dataset.}
    \label{tab:sub-task1-distribution}
\end{table}


\begin{table}[h]
    \small
    \centering
    \begin{tabular}{|c|c|}
        \hline
        \textbf{Classes} & \textbf{Instances}\\
        \hline
        Threats plan to harm, and incitement & 310 \\
        \hline
        Derogation & 1590\\
        \hline
        Animosity & 1165 \\
        \hline
        Prejudiced discussions & 333\\
        \hline
    \end{tabular}
    \caption{Class distribution for sub-task 2 train dataset.}
    \label{tab:sub-task2-distribution}
\end{table}

\begin{table*}[h]
\small
    \centering
    \begin{tabular}{|c|c|}
        \hline
        \textbf{Classes} & \textbf{Instances}\\
        \hline
        Threats of harm & 56 \\
        \hline
        Incitement and encouragement of harm & 254\\
        \hline
        Descriptive attacks & 717 \\
        \hline
        Aggressive and emotive attacks & 673\\
        \hline
        Dehumanising attacks \& overt sexual objectification & 200 \\
        \hline
        Casual use of gendered slurs, profanities, and insults & 637 \\
        \hline
        Immutable gender differences and gender stereotypes & 417\\
        \hline
        Backhanded gendered compliments & 64\\
        \hline
        Condescending explanations or unwelcome advice & 47\\
        \hline
        Supporting mistreatment of individual women & 75\\
        \hline
        Supporting systemic discrimination against women as a group & 258\\
        \hline
    \end{tabular}
    \caption{Class distribution for sub-task 3 train dataset.}
    \label{tab:sub-task3-distribution}
\end{table*}

\section{System Overview}

We have used Transformers with several variations. Based on this variation we try to come up with the best model. 

\begin{figure*}[h]
    \centering
    \includegraphics[scale=0.55]{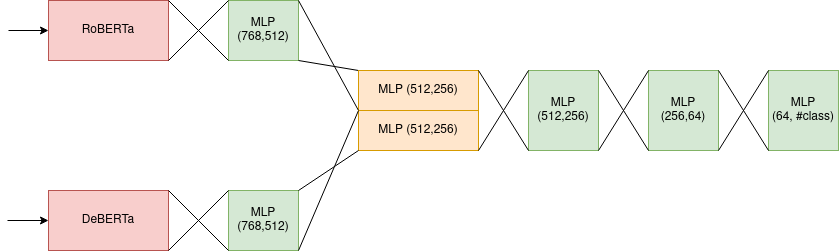}
    \caption{Architecture of our system.}
    \label{fig:Arch}
\end{figure*}

\begin{itemize}
    \itemsep 0em
   \item \textbf{RoBERTa:} It is built on BERT masked language strategy, but excludes BERT’s next-sentence pretraining objective, and training with much larger mini-batches and learning rates. RoBERTa BASE model consists of 12 transformer layers, 12 self-attention heads per layer, and a hidden size of 768. 

    \item \textbf{DeBERTa:} We have used the BASE model which consists of 12 transformer layers, 12 self-attention heads per layer, and a hidden size of 768. It tries to improve RoBERTa by using two techniques: a disentangled attention mechanism and an enhanced mask decoder. 

    \item \textbf{Domain Adaptive Pretraining:} We try to tailor a pretrained model to the domain of a target task. To do so we do masked language modeling over the text related to the domain.
\end{itemize}

With these two transformers, we have experimented with the architecture. The different architecture is as follows:

\begin{enumerate}
    \itemsep 0em
   \item \textbf{RoBERTa/DeBERTa (last layer) + MLP:} We have taken the last layer of the RoBERTa the base model followed by some MLP layers and fine-tuned over the training data. The same setup has been used with DeBERTa as well.  

    \item \textbf{RoBERTa/DeBERTa (average of all layers) + MLP:} Instead of taking the last layer only, here we have taken the average of all the layers in the RoBERTa base model. Which are then followed by some MLP layers and fine-tuned over the training data. The same setup has been used with DeBERTa as well.

    \item \textbf{RoBERTa and DeBERTa concatenation of the last layer + MLP:} Output from the last layers of RoBERTa and DeBERTa were concatenated. Then this is followed by some fully connected layers of MLP. The last layer consists of neurons with a number of classes.

    \item \textbf{RoBERTa and DeBERTa combined (before concatenation of two transformers they are passed through MLP):} Output from the last layers of RoBERTa and DeBERTa were passed through different MLP layers and then concatenated with each other. Then this is followed by some more fully connected layers. The last layer consists of neurons with a number of classes. The model is presented in Fig \ref{fig:Arch}. 

    \item \textbf{Joint Learning for Task B:} For task B, earlier we were only taking the text that is only sexist. But now we will use all the data. And instead of treating the label category which is  class, we are adding another class which implies the text is non-sexist. This makes the problem a 5-class classification problem, rather than a 4-class with the increase in a large amount of data. At the inference time since we are supposed to output only from the 4 classes, we will give the class with the largest probability as output. But if the class is not sexist text, then return the second largest probability class.  

    \item \textbf{Experiment 4 with Domain Adaptive Pretraining:} This model is the same as experiment 4 with the exception that we are using Domain Adaptive Pretrained RoBERTa and DeBERTa. The transformers are trained on the 2M unlabelled texts provided in the task itself.  

\end{enumerate}
All these models are used for all the tasks with a change in the last layer with the number of classes.

\begin{table*}[h]
\small
    \centering
    \begin{tabular}{|c|p{7cm}|c|c|c|c|c|c|}
        \hline
        & & \multicolumn{2}{c|}{\textbf{Task A Macro F1}} & \multicolumn{2}{c|}{\textbf{Task B Macro F1}} & \multicolumn{2}{c|}{\textbf{Task C Macro F1}} \\
        \hline
        \textbf{S.No} & \textbf{Experiment} & \textbf{Val} & \textbf{Test} & \textbf{Val} & \textbf{Test} & \textbf{Val} & \textbf{Test}\\
        \hline
        1 & RoBERTa last layer + MLP & 82.90 & 81.47 & 59.74 & 58.29 & 59.70 & 33.48\\
        \hline
        2 & DeBERTa last layer + MLP & 83.24 & 81.99 & 61.05 & 59.07 & 29.85 & 30.09\\
        \hline
        3 & RoBERTa avg of all layers + MLP & 79.85 & 78.65 & 34.83 & 45.71 & 26.76 & 21.58 \\
        \hline
        4 & DeBERTa avg of all layer + MLP & 80.40 & 78.62 & 59.79 & 55.15 & 22.11 & 21.61 \\
        \hline
        5 & RoBERTa+DeBERTa+ (Embeddings of these two are concatenated before concatenating these are passed through MLP) + MLP & 82.96 & 82.26 & 62.47 & \textbf{62.28} & 33.99 & 31.61 \\
        \hline
        6 & Experiment 5 + Domain Adaptive Pre Training with unlabelled text  & \textbf{84.27} & 82.66 & 63.48 & 60.85 & \textbf{35.30} & \textbf{36.93} \\
        \hline
        7 & Joint Learning for task B using task (A and B’s data), the last layer is of 5 neurons with labels from task B and one non-sexist text & NA & NA & 54.57 & 53.29 & NA & NA \\
        \hline
        8 & RoBERTa+DeBERTa+ (Embedding of RoBERTa and DeBERTa are concatenated and passed through MLP) + Domain Adaptive Pretraining & 84.13 & \textbf{83.9} & \textbf{67.00} & 58.35 & 34.33 & 33.56\\
        \hline
        
    \end{tabular}
    \caption{Results of experiments.}
    \label{tab:expermiment-result}
\end{table*}

\begin{table*}[h]
\small
    \centering
    \begin{tabular}{|p{4cm}|c|c|c|c|}
        \hline
    \textbf{Predicted ->} & Threats, plans to harm and incitement &
Derogation & Animosity & Prejudiced \\

        \hline
        Threats, plans to harm and incitement (Actual) & 62 & 11 & 12 & 4\\
        \hline
        Derogation (Actual)& 16 & 303 & \textbf{107} & 28 \\
        \hline
        Animosity (Actual)& 11 & \textbf{112} & 199 & 11\\
        \hline
        Prejudiced discussions (Actual)& 9 & 27 & 12 & 46\\
        \hline
    \end{tabular}
    \caption{Confusion Matrix for Task B}
    \label{tab:confusionB}
\end{table*}

\section{Experimental setup}

The dataset provided was already divided into train, dev, and test. We have done some preprocessing of the text, which includes converting emoji to text, converting all the characters to lowercase, converting the link to <link> tag, removing punctuations, removing words with numbers, removing stop words, and using WordNet Lemmatizer. But after the cleaning step accuracy was low compared to the non-cleaned text. A probable reason is with transformer tokenization, subword tokenization can perform better than normal tokenization.

The hyperparameter used in this experiment are

epoch = 20, 
Learning Rate = 1e-5,
Loss = Cross Entropy loss,
Optimizer = AdamW.

We have used the Macro F1 score as a measure of evaluation for all three tasks. 

\section{Results}
Results for different experiments are mentioned in Table \ref{tab:expermiment-result}. Our official submission is experiment number 8. Later we experimented with 5 and 6 after the deadline. We have achieved a rank of 35\textsuperscript{th} in task A, 50\textsuperscript{th} in task B and 44\textsuperscript{th} in task C. 

With Domain Adaptive Pretraining the evaluation loss of Roberta changes from 2.49 to 1.98. For Deberta it changes from 11.52 to 2.27. We have also seen reduced perplexity scores, which measure the amount of randomness in our model or how confused the model is, so the less the better. 

\section{Error Analysis}

These error analyses are performed on experiment no 6. 

\subsection{Task A}

\begin{table}[h]
\small
    \centering
    \begin{tabular}{|c|c|c|}
        \hline
         \textbf{Predicted->} & Not sexist & Sexist\\
        \hline
        Not sexist (Actual) & 2909 & 121 \\
        \hline
        Sexist (Actual) & \textbf{346} & 624\\
        \hline
    \end{tabular}
    \caption{Confusion Matrix for Task A}
    \label{tab:confusionA}
\end{table}

With the given confusion matrix in Table \ref{tab:confusionA} test data, we can infer that the model will predict the right label 96\% of the time if the text is not sexist. Issues come when the tweet is sexist. Which is a major problem. Approximately 35\% of the time the model is classifying sexist text as non-sexist, which is a dangerous thing. 

For example, this text: “Don't run away like a little jewish girl.”  and “woman here Opinion discarded.” are treated as not sexist., but they are sexist.

Out of the 346 texts that are sexist but classified as not sexist. Applying some sort of rule-based system where if this word comes then it will be sexist will not work. Because we need to understand the whole context. Most of the time there is no hidden meaning in the sentence. 

One of the main reasons for poor performance in predicting text is sexist, which is less number of samples that are sexist. If we can get more samples that are sexist, our model can try to learn the pattern among them and perform better. We should have tried using distil version of these transformers, as they have less number of weights, which can be trained with fewer samples. And the different distil versions of the transformer can be used, as the number of samples to train on keeps on getting smaller. 

\subsection{Task B}

The confusion matrix for task B is provided in Table \ref{tab:confusionB}. 
The model is confused many times between Derogation and Animosity. Approximately 25\% of the time when the text is derogation it is predicting animosity. And nearly 33\% time when the text is animosity it is treating it as derogation. So the model is not able to clearly distinguish between these classes. Adding more data can help us solve this issue. 

\section{Conclusion}
Domain Adaptive Pre-Training helps us get some extra accuracy numbers in all three sub-tasks. Combining RoBERTa and DeBERTa makes the model big but also brings more variety of aspects to the model. 

Although the macro F1 accuracy is quite good, the data instances where it is failing are very costly in the case of task A. Taking a text down from social media for some time even if it is not sexist is less harmful than keeping a sexist text on social platforms. The model is failing to predict sexist texts. A manual inspection of the failed text of this type suggested that there is a large scope for improvement in the model. 

Despite that, the model shows very promising results, where the number of samples to train on is large. To further improve the accuracy we can use more annotated data from people. Although this will be costly, it will be a one-time effort. 

The model that we submitted for Task B was not able to generalize well. The macro F1 dropped from 67\% to 58.35\%. This can be due to over-fitting or dissimilarity between the distribution of the test and train set. 

One of the ideas that we look forward to working with is to predict pseudo labels for unlabelled tasks. For example, if our best model is sure that the label belongs to a particular task with more than 99\% probability then use that data instance as training data for the model. In this way, we can increase the data to train for our model. 

Another idea that can be looked for is to generate text of a similar type \cite{bhatt-shrivastava-2022-tesla} using decoder-based transformers. For example, we can take text which is less in number and can train on causal language objectives over the labeled samples. This results in more data samples to train on. 

\section{Acknowledgments}
I want to thank Sagar Joshi for helping me understand the task. I would also like to thank Tathagata Raha and Vijayasaradhi Indurthi for giving possible directions to explore the problem. We also want to thank the organizers of this task

\bibliography{anthology,custom}
\bibliographystyle{acl_natbib}

\end{document}